\documentclass{article}

\usepackage{microtype}
\usepackage{graphicx}
\usepackage{subcaption}
\usepackage{booktabs} 
\usepackage{multirow,array,makecell}
\usepackage{float}
\usepackage{amsthm}
\usepackage[htt]{hyphenat}

\usepackage{soul}    

\usepackage{hyperref}

\usepackage[accepted]{icml2021}

\usepackage{balance}
\usepackage{amsmath}
\usepackage{bm}
\usepackage{amssymb}
\usepackage{xcolor}
\usepackage{color}
\usepackage{wrapfig}
\usepackage{adjustbox}
\newcommand{\norm}[1]{\left\lVert#1\right\rVert}

\newcommand\eg{\textit{e.g.}}

\icmltitlerunning{
HoughCL: Finding Better Positive Pairs in Dense Self-supervised Learning
}

\begin{document}

\twocolumn[
\icmltitle{HoughCL: Finding Better Positive Pairs in Dense Self-supervised Learning}




\begin{icmlauthorlist}
\icmlauthor{Yunsung Lee}{sc,ko,naintern}
\icmlauthor{Teakgyu Hong}{na}
\icmlauthor{Han-Cheol Cho}{na}
\icmlauthor{Junbum Cha}{na}
\icmlauthor{Seungryong Kim}{ko}
\end{icmlauthorlist}

\icmlaffiliation{ko}{Korea University}
\icmlaffiliation{sc}{Scatter Lab}
\icmlaffiliation{na}{NAVER Clova AI Research}
\icmlaffiliation{naintern}{Work done during an internship at NAVER Clova}

\icmlcorrespondingauthor{Yunsung Lee}{swack9751@korea.ac.kr}

\icmlkeywords{Machine Learning, ICML}

\vskip 0.3in
]



\printAffiliationsAndNotice{}  

\begin{abstract}

Recently, self-supervised methods show remarkable achievements in image-level representation learning. Nevertheless, their image-level self-supervisions lead the learned representation to sub-optimal for dense prediction tasks, such as object detection, instance segmentation, etc.
To tackle this issue, several recent self-supervised learning methods have extended image-level single embedding to pixel-level dense embeddings. Unlike image-level representation learning, due to the spatial deformation of augmentation, it is difficult to sample pixel-level positive pairs. Previous studies have sampled pixel-level positive pairs using the winner-takes-all among similarity or thresholding warped distance between dense embeddings. However, these naive methods can be struggled by background clutter and outliers problems. In this paper, we introduce Hough Contrastive Learning (HoughCL), a Hough space based method that enforces geometric consistency between two dense features. 
HoughCL achieves robustness against background clutter and outliers. Furthermore, compared to baseline, our dense positive pairing method has no additional learnable parameters and has a small extra computation cost. 
Compared to previous works, our method shows better or comparable performance on dense prediction fine-tuning tasks.


\end{abstract}


\section{Introduction}
\label{s_introduction}
 
Recent self-supervised visual representation learning methods have made significant progress in image recognition since InfoNCE~\cite{oord2018representation} based contrastive representation learning. Most of recent self-supervised visual representation learning~\cite{chen2020simple,he2020momentum,chen2020improved,caron2018deep,NEURIPS2020_swav,NEURIPS2020_byol,chen2021exploring,ermolov2020whitening,zbontar2021barlow,bardes2021vicreg} have in common with maximizing the agreement between positive pairs of embedding vectors that are sampled from different views of the same image.
However, most of these self-supervised methods only consider image-level embeddings lacking local information. It may be appropriate for image-level recognition tasks but can be sub-optimal for dense prediction tasks such as object detection or semantic segmentation.

Several recent works~\cite{wang2021dense,pixpro,roh2021spatially} have learned representations from pixel-level densely embedded vectors, and show improvements when transferring to downstream dense prediction tasks. In image-level self-supervised learning, positive pairs were easily assigned, because image-level features are invariant in data augmentation. However, since pixel-level features are variant in augmentation, it is difficult to assign pixel-level positive pairs. In DenseCL~\cite{wang2021dense}, they introduce a dense projection head that outputs dense feature vectors. To obtain positive pixel pairs, they simply calculate the cosine similarity between pixel vectors and choose the positive pair which has the highest similarity value. This simple winner-takes-all method suffers from background clutter and outliers. Meanwhile in PixPro~\cite{pixpro}, in addition to the dense projection head, an asymmetric network is introduced that computes its smoothed transform by propagating pixel-level features. Their assignments of dense positive pairs differ from DenseCL. Each point in a feature map is first warped to the original image space, and the distances between all pairs of points from the two feature maps are computed. By thresholding these distances, they assigned dense positive pairs. Though this method can simply find a positive pixel pair, there is a risk of semantically different pixels paired as positive because all pixels located close to each other are treated as positive. 
In addition, the threshold value is a hyper-parameter that requires a new manual setting.

In this paper, we introduce the pixel-level dense positive pairing method based on Hough geometric voting, inspired by the algorithm of~\citet{cho2015unsupervised}. Through weighted voting in Hough space, we can obtain geometrical consistent dense positive pairs. This geometric consistency gives a model robustness against background clutter and outliers. Furthermore, it does not require additional training parameters. Thus, our method is generally applicable to self-supervised learning methods where matching of dense positive pairs exists. Compared to previous works, our method shows better or comparable performance on dense prediction fine-tuning tasks. In particular, experimental results of pre-training on Tiny ImageNet, a miniature of ImageNet, our method outperforms the DenseCL when transferring to downstream dense prediction tasks, including PASCAL VOC object detection (+3.0\% AP), COCO object detection (+1.1\% AP) and COCO instance segmentation (+0.9\% AP). 



\begin{figure*}[t]
\centering
\captionsetup[subfigure]{aboveskip=0pt,belowskip=0pt}
\begin{subfigure}[b]{0.36\textwidth}
    \includegraphics[width=\textwidth]{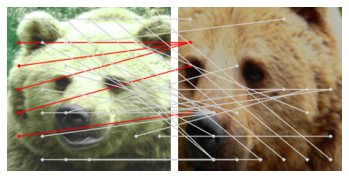}
\end{subfigure}
\begin{subfigure}[b]{0.36\textwidth}
    \includegraphics[width=\textwidth]{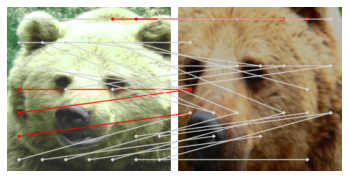}
\end{subfigure}
\\
\begin{subfigure}[b]{0.36\textwidth}
    \includegraphics[width=\textwidth]{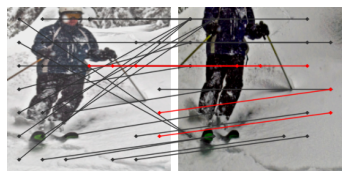}
    \caption{\textit{Dense Positive Pairs in DenseCL}}
\end{subfigure}
\begin{subfigure}[b]{0.36\textwidth}
    \includegraphics[width=\textwidth]{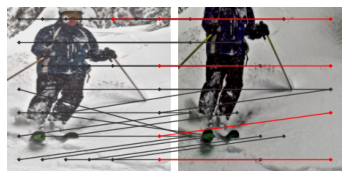}
    \caption{\textit{Dense Positive Pairs in HoughCL}}
\end{subfigure}

\vspace{-0.5em}
\caption{\textbf{Visualization of dense positive pairs in DenseCL~\cite{wang2021dense} and our HoughCL} Both methods are pre-trained 800 epochs on the COCO dataset and have a ResNet-50 backbone network. The red line segment represents the five pairs with the highest confidence, and the gray line segment represents the 20 pairs with the lowest confidence. The dense positive pairs of HoughCL are geometrical consistent and robust against background clutter and outliers compared with DenseCL.}
\vspace{-1.0em}
\label{fig:whyrhm}
\end{figure*}

\section{Background}
\label{s_background}

After self-supervised learning has become a new paradigm for pre-training in image-level recognition tasks, several recent self-supervised learning methods are designed for dense prediction tasks. DenseCL~\cite{wang2021dense} introduces a dense projection head that outputs dense feature vectors, and it is a simple but effective way to learn a pixel-level dense representation. In this section, we briefly review our baseline method, DenseCL.

\subsection{DenseCL: Dense Contrastive Learning}
\label{sub_densecl}

Compared to the existing paradigm, the core differences of DenseCL lie in the encoder and loss function.
Given an input view, the dense feature maps are extracted by the backbone network, \eg, ResNet~\cite{he2016deep}, and forwarded to the following projection head. The projection head consists of two sub-heads in parallel, which are global projection head and dense projection head, respectively. The global projection head can be any of the existing projection heads in \citet{he2020momentum,chen2020simple,chen2020improved}, which takes the dense feature maps as input and generates a global feature vector for each view. In contrast, the dense projection head takes the same input but outputs dense feature vectors. The backbone and two parallel projection heads are end-to-end trained by optimizing a joint pairwise loss at the levels of both global features and local features.

In DenseCL, dense contrastive loss is extending the original contrastive loss function to a dense paradigm.
$\{ t_0, t_1, ... \}$ is a set of encoded keys for each encoded query $r$.
$r$ corresponds to one of the $S_h \times S_w$ (for simpler illustration, they use $S_h=S_w=S$) feature vectors generated by the dense projection head.
Each negative key $t_-$ is the pooled feature vector of a view from a different image.
The positive key $ t_+$ is assigned according to the extracted correspondence across views, which is one of the $S^2$ feature vectors from another view of the same image.
The dense contrastive loss is defined as: 
$\mathcal{L}_{r} = \frac{1}{S^2} \sum_{s} -\log \frac{\exp(r^s {\cdot} t^s_{+} / \tau)} {\exp(r^s {\cdot} t^s_{+} / \tau) + \sum_{t^s_-} \exp(r^s {\cdot} t^s_-  / \tau)},$


where $r^{s}$ denotes the $s^{\rm{th}}$ out of $S^2$ encoded queries. 

Overall, the total loss for DenseCL can be formulated as $\mathcal{L} = (1 - \lambda)\mathcal{L}_{q} + \lambda \mathcal{L}_{r}$, where $\mathcal{L}_{q}$ is existing image-level contrastive loss~\cite{oord2018representation}. 
$\lambda$ is set to 0.5 which is validated by experiments in  \citet{wang2021dense}.

\subsection{Dense Positive Pairs in DenseCL}
\label{sub_correspondence}

\def\SM{\mathbf{\Delta}}
\def\vd{\mathbf{d}}
\def\mF{\mathbf{F}}
\def\mD{\mathbf{\Theta}}

In DenseCL, the dense correspondence between the two views of the same input image is the dense positive pairs $ t_+$ described in \ref{sub_densecl}.
For each view, the backbone network extracts feature maps $\mF \in \mathbb{ R}^{H\times W\times K}$, and the dense projection head extracts dense feature vectors $ \mD  \in \mathbb{ R}^{ S_h \times  S_w \times E}$.
The correspondence between the dense feature vectors from the two views, $\mD_1$ and $ \mD_2$, is made using the backbone feature maps $\mF_1$ and $\mF_2$.
The $\mF_1$ and $\mF_2$ are first downsampled to have the spatial shape of $S \times S$ by an adaptive average pooling, and then used to calculate the cosine similarity matrix $\SM \in \mathbb{ R}^{S^2 \times S^2}$.
The matching rule is winner-takes-all in feature vector similarity.
The matching process can be formulated as $c_i = \mathop{\arg\max}_{j} sim(\bm f_i,\bm f^\prime_j)$.
where  
$\bm f_i$ is the $i^{\rm th}$ feature vector of $\mF_1$, and $\bm f^\prime_j$ is the $j^{\rm th}$  of $\mF_2$, and ${sim}(\bm u,\bm v)$ denotes the cosine similarity.
It means that the positive pair of $i^{\rm th}$ feature vector of $\mD_1$ is ${c_i}^{\rm th}$ of $\mD_2$.

\section{Hough Contrastive Learning}
\label{s_method}
The Hough transform~\cite{hough1962method} is a classic method developed to identify primitive shapes in an image via geometric voting in a parameter space.
In geometric matching, \citet{cho2015unsupervised} first extends it to the Probabilistic Hough Matching (PHM) algorithm for unsupervised object discovery.
Recent semantic alignment and correspondence methods~\cite{min2019hyperpixel,min2020learning,liu2020semantic,min2021convolutional} employ Hough matching. Through Hough matching, these methods can form matches considering geometric consistency as well as appearance similarity.

As can be seen in~\ref{fig:whyrhm}, previous  dense positive sample pairing methods, \eg, \texttt{argmax} in DenseCL~\cite{wang2021dense} and thresholding warped distance in PixPro~\cite{pixpro}, may suffer from background clutter and outliers. These mismatches can give poor guidance information for the model to learn dense representations. 
To give the model robust guidance information against background clutter and outliers, we introduce geometric consistent dense positive pairing with PHM.
The key idea of PHM is to re-weight appearance similarity by Hough space voting to enforce geometric consistency. By applying the PHM principle, we propose dense positive matching method that maintains more geometrical tendencies. In this paper, our method was applied to the baseline DenseCL, but it can be generally applied to self-supervised learning methods that use dense positive pairs. 

In our context, let $\mathcal{D}=(\mathcal{H}, \mathcal{H}')$ be two sets of dense projected features, and $m=(\mathbf{h},\mathbf{h}')$ be a region vector match where $\mathbf{h}$ and $\mathbf{h}'$ are respectively elements of $\mathcal{H}$ and $\mathcal{H}'$. Given a Hough space $\mathcal{X}$ of possible offsets (image transformation) between the two dense projected features,  
the confidence for match $m$, $p(m|\mathcal{D})$, is computed as  
\begin{align}
    \vspace{-3em}
    \label{eqn:houghmatching} 
    p(m|\mathcal{D}) &\propto  p(m_\mathrm{a})\sum_{\mathbf{x}\in \mathcal{X}}p(m_\mathrm{g}|\mathbf{x})\sum_{m \in \mathcal{H} \times \mathcal{H}'}p(m_\mathrm{a})p(m_\mathrm{g}|\mathbf{x}),  
    \vspace{-3em}
\end{align}
where $p(m_\mathrm{a})$ represents the confidence for similarity matching and $p(m_\mathrm{g}|\mathbf{x})$ is the confidence for geometric matching with an offset $\mathbf{x}$, measuring how close the offset induced by $m$ is to $\mathbf{x}$, and implemented by a discretized Gaussian kernel centered on $\mathbf{x}$. 
By sharing the Hough space $\mathcal{X}$ for all matches, PHM efficiently computes the match confidence.
Matching confidence is computed as the exponential cosine similarity, $p(m_\mathrm{a})
    = \mathrm{ReLU}\Big( \frac{\mathbf{f} \cdot \mathbf{f}'}{\norm{\mathbf{f}} \norm{\mathbf{f}'}} \Big)^d$.
The ReLU function clamps negative values to zero and the exponent $d \geq 2$ improves matching performance by suppressing noisy activations. We set $d=3$ in our experiments.

Following the strategy of \citet{min2019hyperpixel} to compute $p(m_\mathrm{g}|\mathbf{x})$, we construct a two-dimensional offset space, quantize it into a grid of bins, and use a set of center points of the bins for $\mathcal{X}$. For Hough voting, each match $m$ is assigned to the corresponding offset bin to increment the score of the bin by the appearance similarity score, $p(m_{\mathrm{a}})$. Despite their (serial) complexity of $O(|\mathcal{H}| \times |\mathcal{H}'|)$, the operations are mutually independent, and can thus easily be parallelized on a GPU. In Tiny-ImageNet pre-training, DenseCL took 1'28'' and HoughCL took 1'34'' time per epoch (with 8 V-100 GPU machine). The overhead is less than 6\%.

\section{Experiments}
\label{s_experiments}


\subsection{Pre-training Setup}


To validate the performance of our method on various datasets, we conduct experiments on not only COCO and ImageNet, which are mainly used in the other methods, but also Tiny ImageNet, which is a relatively small dataset.
COCO~\cite{lin2014microsoft} consists of about 118K training images which containing common objects in complex everyday scenes.
ImageNet~\cite{deng2009imagenet} consists of about 1.28M training images in 1K image classes.
Tiny ImageNet~\cite{le2015tiny} is a miniature of ImageNet.
It consists of 100K training images of size 64$\times$64 in 200 image classes.

The pre-training setup mostly follows DenseCL~\cite{wang2021dense}.
A ResNet-50~\cite{he2016deep} is adopted as a backbone.
SGD optimizer is utilized and its weight decay and momentum are set to 0.0001 and 0.9, respectively.
The initial learning rates are set to 0.5, 0.3, and 0.03 in Tiny ImageNet, COCO, and ImageNet, respectively and cosine annealing schedule is used.
The batch size is set to 256, using 8 V100 GPUs.
The number of training epochs are set to 200, 800, and 200 in Tiny ImageNet, COCO, and ImageNet, respectively.

\subsection{Fine-tuning Setup}

We evalute the pre-trained model on three downstream dense prediction tasks: PASCAL VOC object detection~\cite{everingham2010pascal}, and COCO object detection and instance segmentation~\cite{lin2014microsoft}.
The fine-tuning setup follows DenseCL.

\subsection{Results}

\begin{table}[t]
    \caption{Experimental results of PASCAL VOC object detection. A Faster R-CNN (C4-backbone) is fine-tuned on \texttt{trainval07+12} set for 24K iterations and evaluated on \texttt{test2007} set. The results are averaged over 2 independent trials. $^\dagger$ indicates the scores are reported from \cite{wang2021dense}}
    \label{tbl:voc_detection}
    \vskip 0.1in
    \centering
    \begin{tabular}{l|l|ccc}
         Dataset & Method & AP & AP$_{\text{50}}$ & AP$_{\text{75}}$ \\ \toprule
         -              & random init.$^\dagger$ & 32.8 & 59.0 & 31.6 \\ \midrule
         Tiny           & MoCo v2 & 47.6 & 75.3 & 51.2 \\
         ImageNet       & DenseCL & 47.5 & 74.6 & 51.2 \\
                        & HoughCL & \textbf{50.5} & \textbf{76.9} & \textbf{55.0} \\ \midrule
         COCO           & MoCo v2$^\dagger$ & 54.7 & 81.0 & 60.6 \\
                        & DenseCL$^\dagger$ & 56.7 & 81.7 & \textbf{63.0} \\
                        & HoughCL & \textbf{56.8} & \textbf{82.1} & \textbf{63.0} \\ \midrule
         ImageNet       & super. IN$^\dagger$ & 54.2 & 81.6 & 59.8 \\
                        & MoCo v2$^\dagger$ & 57.0 & 82.2 & 63.4 \\
                        & DenseCL$^\dagger$ & \textbf{58.7} & \textbf{82.8} & 65.2 \\
                        & HoughCL & 58.5 & 82.6 & \textbf{65.7} \\ \bottomrule
    \end{tabular}
    \vskip -0.1in
\end{table}

Table~\ref{tbl:voc_detection} shows the experimental results of PASCAL VOC object detection.
HoughCL outperforms the other methods in Tiny ImageNet and shows similar performance when pre-trained on COCO and ImageNet.
In Tiny ImageNet, HoughCL achieves 3.0\% AP and 3.8\% AP$_{\text{75}}$ improvements compared to DenseCL.
This result indicates the efficiency of HoughCL by showing superior performance in relatively small scale dataset.
In COCO and ImageNet, HoughCL shows similar AP scores compared to DenseCL, but it achieves slightly better AP$_{\text{75}}$ scores in ImageNet.

\begin{table}[h]
    \caption{Experimental results of COCO object detection. A Mask R-CNN detector (FPN backbone) is fine-tuning on \texttt{train2017} split with 1$\times$ schedule and evaluated on \texttt{val2017} split. The results are averaged over 2 independent trials. $^\dagger$ indicates the scores are reported from \cite{wang2021dense}}
    \label{tbl:coco_detection}
    \vskip 0.1in
    \centering
    \begin{tabular}{l|l|ccc}
         Dataset & Method & AP & AP$_{\text{50}}$ & AP$_{\text{75}}$ \\ \toprule
         -              & random init.$^\dagger$ & 32.8 & 50.9 & 35.3 \\ \midrule
         Tiny           & MoCo v2 & 35.6 & 54.6 & 38.8 \\
         ImageNet       & DenseCL & 35.4 & 54.0 & 38.6 \\
                        & HoughCL & \textbf{36.5} & \textbf{55.4} & \textbf{39.9} \\ \midrule
         COCO           & MoCo v2$^\dagger$ & 38.5 & 58.1 & 42.1 \\
                        & DenseCL$^\dagger$ & \textbf{39.6} & \textbf{59.3} & \textbf{43.3} \\
                        & HoughCL & 39.5 & \textbf{59.3} & 43.1 \\ \midrule
         ImageNet       & super. IN$^\dagger$ & 39.7 & 59.5 & 43.3 \\
                        & MoCo v2$^\dagger$ & 39.8 & 59.8 & 43.6 \\
                        & DenseCL$^\dagger$ & \textbf{40.3} & \textbf{59.9} & \textbf{44.3} \\
                        & HoughCL & 40.0 & \textbf{59.9} & 43.6 \\  \bottomrule
    \end{tabular}
    \vskip -0.1in
\end{table}

\begin{table}[h]
    \caption{Experimental results of COCO instance segmentation. A Mask R-CNN detector (FPN backbone) is fine-tuning on \texttt{train2017} split with 1$\times$ schedule and evaluated on \texttt{val2017} split. The results are averaged over 2 independent trials. $^\dagger$ indicates the scores are reported from \cite{wang2021dense}}
    \label{tbl:coco_segmentation}
    \vskip 0.1in
    \centering
    \begin{tabular}{l|l|ccc}
         Dataset & Method & AP & AP$_{\text{50}}$ & AP$_{\text{75}}$ \\ \toprule
         -              & random init.$^\dagger$ & 29.9 & 47.9 & 32.0 \\ \midrule
         Tiny           & MoCo v2 & 32.5 & 51.8 & 34.9 \\
         ImageNet       & DenseCL & 32.2 & 51.3 & 34.5 \\
                        & HoughCL & \textbf{33.1} & \textbf{52.6} & \textbf{35.5} \\ \midrule
         COCO           & MoCo v2$^\dagger$ & 34.8 & 55.3 & 37.3 \\
                        & DenseCL$^\dagger$ & \textbf{35.7} & \textbf{56.5} & \textbf{38.4} \\
                        & HoughCL & \textbf{35.7} & 56.4 & 38.2 \\ \midrule
         ImageNet       & super. IN$^\dagger$ & 35.9 & 56.6 & 38.6 \\
                        & MoCo v2$^\dagger$ & 36.1 & 56.9 & 38.7 \\
                        & DenseCL$^\dagger$ & \textbf{36.4} & \textbf{57.0} & \textbf{39.2} \\
                        & HoughCL & 36.2 & 56.8 & 38.8 \\ \bottomrule
    \end{tabular}
    \vskip -0.1in
\end{table}

Table~\ref{tbl:coco_detection} and Table~\ref{tbl:coco_segmentation} show the experimental results of COCO object detection and instance segmentation.
Similar to the results of PASCAL VOC, HoughCL shows superior performances in Tiny ImageNet.
It outperforms DenseCL by 1.1\% AP and 1.3\% AP$_{\text{75}}$ in object detection and 0.9\% AP and 1.0\% AP$_{\text{75}}$ in instance segmentation.
In COCO and ImageNet, HoughCL achieves similar or slightly lower scores than other methods.


Overall, HoughCL shows superior performance in Tiny ImageNet, but similar performance in COCO and ImageNet.
We think this is because HoughCL is not yet optimized for COCO and ImageNet.

\section{Conclusion}
\label{s_conclusion}
In this paper, we introduce a novel dense positive sample pairing method based on Hough geometric voting. 
Proposed method provides geometrically consistent dense positive pairs through weighted voting in Hough space. This geometric consistency gives a model robustness against background clutter and outliers. 
Experimental results show that HoughCL outperforms baselines especially in Tiny ImageNet, which consists of downsampled images from ImageNet. It empirically demonstrates HoughCL matches dense positive pair robustly in noisy setting, \eg, downsampling noise.

On the other hand, our method performs similar to the baselines on ImageNet and COCO datasets. Although this work improves robustness in dense representation learning, we believe that the geometrical consistency has the potential to improve the performance even on less noisy datasets, such as ImageNet and COCO. We will cover this topic in the future work.


\clearpage
\bibliography{6.reference}
\bibliographystyle{icml2021}



\end{document}